\documentclass[11pt]{article}
\usepackage{graphicx}
\usepackage{color}
\usepackage[colorlinks,allcolors=blue]{hyperref}
\usepackage{url}
\usepackage[authoryear,round]{natbib}
\setcitestyle{round,aysep={},yysep={;}}
\usepackage{a4wide}

\renewcommand{\cite}{\citep}
\newcommand*\rot{\rotatebox{90}}

\author{Saturnino Luz} \title{Computational linguistics and Natural
  Language Processing\footnote{This is the unedited author's copy of a
    text which appeared as a chapter in "The Routledge Handbook of
    Translation and Methodology'', edited by F Zanettin and C Rundle
    (2022), \url{https://doi.org/10.4324/9781315158945}}}

\begin{document}

\maketitle


This chapter provides an introduction to computational linguistics
methods, with focus on their applications to the practice and study of
translation. It covers computational models, methods and tools for
collection, storage, indexing and analysis of linguistic data in the
context of translation, and discusses the main methodological issues
and challenges in this field.  While an exhaustive review of existing
computational linguistics methods and tools is beyond the scope of
this chapter, we describe the most representative approaches, and
illustrate them with descriptions of typical applications. 

\section{Introduction, definitions}
\label{sec:introduction}

Broadly defined, the term {\em computational linguistics} refers to
the use of computational methods and tools in the study of linguistic
phenomena. A distinction is sometimes made between computational
linguistics and {\em natural language processing}. The former is
usually regarded as the study of linguistic ability as a computational
process, and the latter as an ``engineering'' pursuit directed towards
the application of algorithmic methods to practical problems such as
automatic categorisation of text, parsing, prediction of the
part-of-speech (categories) of words, automatic translation, text
summarisation and other tasks, all of which involve processing of
natural languages\footnote{\citet[p. 1]{bib:GliozzoStrapparava09sdcl}
  attribute the distinction between computational linguistics and
  natural language processing to the late Martin Kay.} (as opposed to
formal or programming languages). As the boundaries between
computational linguistics and natural language processing are not
always clearly defined, we will not concern ourselves with this
distinction here. Therefore, our use of the term computational
linguistics in this chapter will encompass the methods and tools these
fields share in common, with a focus on methodology and its relation
to translation practice and enquiry.

As computing systems have become increasingly important in the practice of
translation, their use in scholarly studies of translation and its
practice have followed a similar path. A typical example of this trend
is the consolidation of corpus-based translation studies as an active
area of research \citep{bib:Bakerothers93c}. As with all data
intensive, corpus-based linguistic studies, the applications of corpora
in the area of translation studies owe their existence to
computational tools and resources, mirroring the evolution of
translation practice \cite{bib:KaramanisLuzDoherty11tpw}.

While computational linguistics is usually associated with
quantitative or formal approaches, its methods have also been used to
support qualitative analysis. In fact, it can be argued that the
applications of computational linguistics to corpus analysis, and
corpus-based translation studies in particular, are generally based on
an interplay of qualitative and quantitative techniques
\cite{bib:McEneryHardie11cl}. The combined use of concordancing (a
qualitative method) and word frequency analysis (a quantitative
method) provides a well known example of such interplay. However, more
intricate methods and combinations exist and continue to appear as the
field of computational linguistics evolves. In the following section,
we review the main developments in the field, the tools
they produced, and the uses these tools have found in translation and
methodology.

Before we proceed, however, a few basic definitions are in order. In
addition to the central concept of computational linguistics, we will
often refer to related concepts such a corpus, metadata, corpus
linguistics, corpus based translation studies, statistical methods,
machine learning, and text visualisation. A {\em corpus}, in this
context, is simply a collection of texts often accompanied by data
that describe and categorise the texts in the collection, commonly
referred to as {\em metadata}, and sometimes complemented by resources
of a non-textual nature, such as images, recorded speech and video. We
will limit ourselves here to corpora comprising text and metadata
only. {\em Corpus linguistics}, sometimes referred to {\em
  data-intensive linguistics} \cite{bib:ChurchMercer93in}, is the use
of corpora, in digital form, aided by computing technology, for the
study of linguistic phenomena. {\em Corpus based translation studies}
employs and extends corpus linguistics methods in the study of
translated text \cite{bib:Laviosa04c} with the aim of uncovering the
particular ``nature of translated text as a mediated communicative
event'' \cite{bib:Bakerothers93c}. The data-intensive nature of corpus
linguistics implies a preponderance of {\em statistical methods},
which encompass both descriptive (e.g. word frequency, co-occurence
counts) and inferential statistics (e.g. inference of collocation
patterns from co-occurence probabilities). Traditional statistical
methods are increasingly being complemented by {\em machine learning}
methods. Machine learning is a sub-discipline of artificial
intelligence which aims to automate the processes of creating
structured representations from raw data (e.g. representing text as
feature vectors) and performing inference on these representations
(e.g. classifying text into high-level categories, labelling a
sequence of words with the corresponding sequence of grammatical
categories, parsing sentences into tree structures). Finally, the
analytic framework of corpus linguistics, and corpus based translation
studies, includes descriptive tools that enhance the ability of the
translation scholar to inspect large volumes of text data iteratively,
with the help of a visual computer interface. These tools will be
referred to as {\em text visualisation} tools, and include the
familiar concordance list, as well as a number of new visual presentations
that have been developed more recently.

\section{Methods, Tools and Users}
\label{sec:participants}

Computational linguistic methods can be categorised as {\em symbolic
  methods} and {\em statistical learning methods}. The former
consists of approaches that originated from the formal logic and
symbolic artificial intelligence (AI) tradition. The latter have their
roots in probability theory, statistics and connectionist (neural
network) approaches.
While handcrafted symbolic parsers and machine translation systems
have all but disappeared, symbolic approaches still play a role in an
area that is relevant to translation, namely, semantics, where large
scale ontologies and lexical databases such as Wordnet
\cite{bib:Miller95w} have been employed in corpus linguistics
\cite{bib:BudanitskyHirst06evwml} as well as in knowledge based machine
translation systems \cite{bib:Costa-JussaFonollosa15l}.

However, it is statistical methods that currently dominate the
contributions of computational linguistics to translation methodology,
studies, and practice. The above mentioned techniques of concordancing and frequency
list comparison can be regarded as statistical methods. Relative
frequencies are clearly statistical in nature, and text concordances
are a form of descriptive statistics combined with a text
visualisation device, as can be seen clearly in the abstract
representations proposed by
\citet{bib:LuzSheehanAVI14,bib:WattenbergViegas08}, for instance. In
addition to these basic methods, other statistical measures have long
been used in corpus linguistics \cite{bib:Sinclair91CorpusConcordance}
and translation studies.

More recently, machine learning methods have become closely linked to
computational linguistics, and their influence is also evident in
corpus based translation studies. Automatic classification methods
\cite{bib:EmmsLuz07MLFNLP,bib:Sebastiani2002} have been used, for
instance, in the identification of linguistic and stylistic patterns
that might be characteristic of translated text, or ``translationese''
\cite{bib:BaroniBernardini05}. Machine learning has also extended the
repertoire of distributional semantics beyond statistical and
information theoretic measures, providing this area with sophisticated
modelling tools which include latent semantic analysis
\cite{bib:DeerwesterDumaisEtAl90in}, latent Dirichlet allocation for
topic modelling \cite{bib:Blei12p,bib:BleiNgJordan03l}, and word
embedding models such as word2vec \cite{bib:MikolovSutskeverEtAl13d}
and t-distributed stochastic neighbour embedding, or t-SNE
\cite{bib:VanHinton08vhdd}. Machine translation have also experience
great improvements in recent years \cite{bib:WuSchusterEtAl16g} due to
advances in the area of deep learning \cite{bib:LeCunBengioHinton15d}.

Underlying most of these methods, is an infrastructure for
pre-processing the corpora to be analysed. This involves 
collecting, storing, indexing and providing access to language
resources. The word-wide web has been used both as a source of such
data \cite{bib:BaroniBernardiniEtAl09w} and as a medium for provision
of access to corpora \cite{bib:LuzCTSBook2011}.   

We will now describe these different facets of computational
linguistics in translation methodology and study, from basic
infrastructure issues and tools to advanced techniques and methods.

\subsection{Building and managing a corpus}

Gathering, storing, indexing and managing access to a collection of
electronic texts are usually the first steps in any type of corpus
based study. A number of tools have been developed over the years to
support these tasks, sometimes individually, as in the case of UNIX
directory structures and command line tools for search and filtering
of text \cite{bib:SchmittChristiansonGupta07lun}, sometimes as part of
larger general-purpose packages such as the indexing tools provided by
the Xapian project\footnote{https://xapian.org/}, and sometimes as
specialised corpus linguistics tools. While reviewing the first two
kinds of tools falls outside the scope of this chapter, and an
exhaustive review of specialised corpus linguistics tools would be
impractical, we describe three tools used for corpus management and
analysis which illustrate the main characteristics of the various
existing tools and their differences.

As regards differences, existing tools can be can be grouped into
broad categories with respect to their mode of distribution, the way
they store and provide access to corpora, and their licensing
terms. The mode of distribution of corpus software can be distribution
through conventional stand-alone software packages, distribution
through web-services and on-line interfaces, or mixed web-based
distribution of specialised software ``clients'' interacting with one
or more ``servers''.  

One of the most popular tool set distributed in the conventional way
is WordSmith tools \cite{bib:Scott21wtvs}. WordSmith tools is a
robust and stable set of tools which has been used in a number of
papers in lexicography and translation studies. However, it only runs
on Windows platforms, which limits its accessibility. Other similar
concordance software such as MonoConc and ParaConc
\cite{bib:Barlow99m} also have this limitation.
Possibly the
best known example of web-based corpus tools is the Sketch Engine
\cite{bib:KilgarriffBaisaEtAl14sen}. This tool, which runs entirely on
web browsers capable of running Javascript has had great impact on
corpus and lexicography research. Finally, an example of a mixed
distribution mode is the modnlp/tec tool \cite{bib:LuzCTSBook2011},
which is distributed using Java Web Start technology and runs as a
stand-alone client on the user's computer, typically interacting with
a remote server. The modnlp software suite runs on most computer
operating systems, and has been used extensively in translation
studies, having been initially developed for the translational English
corpus, TEC \cite{bib:BakerIJCL99}, and having since been used in a
number of other applications, most recently the Genealogies of
Knowledge projects\footnote{http://www.genealogiesofknowledge.net/}
\cite{bib:Jones19ss,bib:LuzSheehan20dgk}.

In terms of the way they provide access to corpora, WordSmith tools
only allow indexing of and access to data that reside on the user's
computer. They are therefore typically used with smaller corpora not
subjected to copyright restrictions. Sketch Engine is used mainly for
access to large, pre-indexed, web-stored corpora, but also allows
users to load and index their own corpus, as well as define new
web-based corpora. The modnlp/tec tool is capable of both accessing
corpora stored on the user's computer, as well as remote web-based
corpora (its original motivation was to provide the research community
with on-line access to material subjected to copyright restrictions in
a way that did not violate copyright).

These tools also differ with respect to the licenses under which they
are distributed. Both WordSmith tools and the Sketch Engine are
commercial proprietary software. WordSmith is distributed under a
proprietary license as executable binary code. No access to source
code is provided. Sketch Engine is fully web based, and operates on a
subscription basis. As in the case of WordSmith tools, no access to
source code is provided with the Sketch Engine software. The modnlp
suite on the other hand is free software \cite{bib:Stallman02fsfs} and
is distributed under the terms of the GNU general public license, a
free/libre and open source software (FLOSS) license.

Common functionality provided by these and most other corpus
management tools include tokenisation, indexing, and data
presentation. Wordsmith tools and modnlp have built-in indexing
capabilities, and the latter can either access a pre-built index via
the network, or create a local index, using different tokenisers
(e.g. lucene, Stanford tokeniser for Arabic, etc) as needed. Sketch
engine only provides access to indexing via its web interface. The
index is otherwise inaccessible to the user. While both Wordsmith
tools and Sketch Engine have basic features for storing metadata,
modnlp provides comprehensive support for combining text search and
metadata constraints. Metadata files in modnlp are encoded in XML
\cite{bib:LuzCTSBook2011} and managed by the eXist-db
database backend\footnote{http://exist-db.org/}, which supports the
XQuery language, as W3C standard. Data presentation minimally
covers concordance, collocation and frequency lists. Presentation
techniques are described below.

\subsection{Data presentation and visualisation}
\label{sec:data-pres-visu}

The concordance list is perhaps the oldest and most basic form of data
presentation for corpus analysis. It consists of arranging of passages
of a text or collection of texts in alphabetical order according to
user-defined keywords. Combined with a ``keyword-in-context'' (KWIC)
indexing method, which aligns different occurrences of the keyword
specified by the user at the centre of each concordance line, and with
interactive capabilities for sorting and rearranging the list,
concordancing is a powerful text presentation method.

\begin{figure}[htbp]
  \centering
  \includegraphics[width=\linewidth]{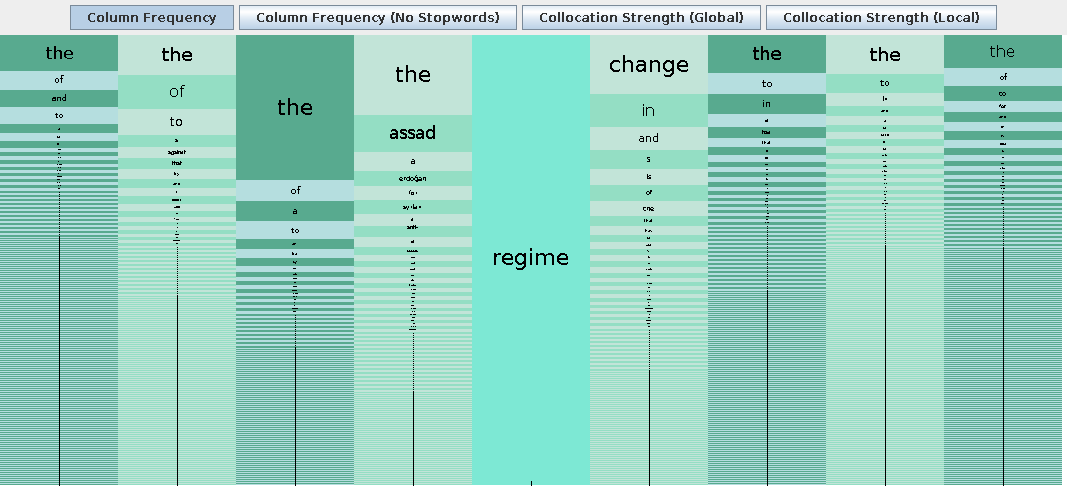}\\[1ex]
  \includegraphics[width=\linewidth]{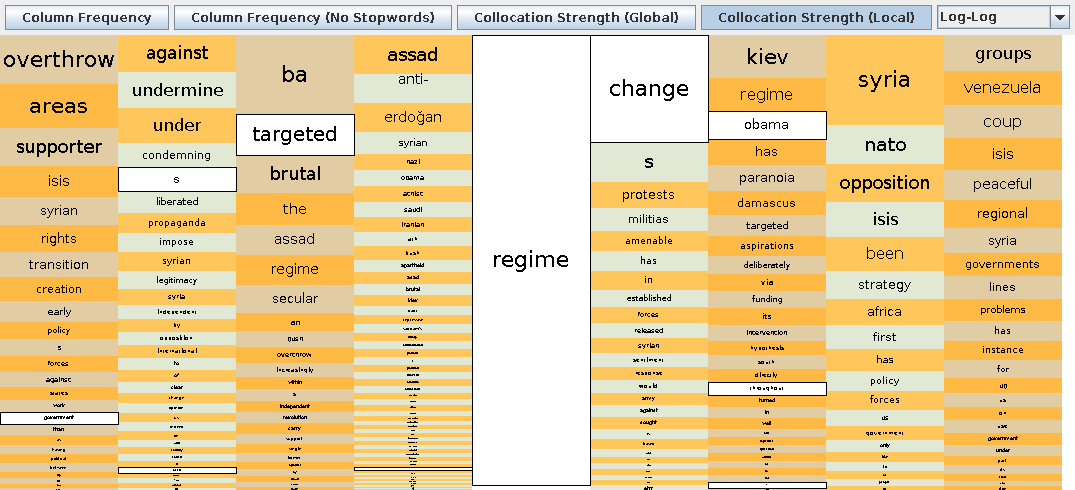}\\
  \caption{Raw frequency distribution view (top) and collocation
    strength view (bottom) for the
    word "regime" in the Genealogies of Knowledge modern corpus. The
    collocation strength view is scaled according to mutual information
    statistics and reveals patterns not readily evident in the
    frequency distribution.}
  \label{fig:mosaic}
\end{figure}

Visual generalisations of concordances have been proposed, which seek
to address one of the main limitations of this technique, namely its
poor use of space, since a concordance could comprise several thousand
lines of text. \citet{bib:WattenbergViegas08} proposed a form of
visual encoding of concordances called Word Tree, which displays
alternatively the left or the right context of a concordance as a tree
where words are vertices linked in textual order and scaled in size
according to their frequencies. Although they provide a better visual
summary than concordance lists, Word Trees limit the display to half
of the text (the keyword plus the left or right context) of its
underlying concordances.  It is therefore impossible for the user to
read the full sentences in which the keyword appears. For certain
corpus linguistics tasks, such as detection of phrases that span left
and right contexts (as in the expression ``run the whole {\em gamut}
of ...'', for instance), frequency information for words occurring on
each context is usually more useful to the analyst than the linear
structure of a single context \cite{bib:Sinclair03Reading}. A mosaic
style visualisation has been proposed which overcomes this limitation
\cite{bib:LuzSheehanAVI14}.  This ``concordance mosaic'' uses a tabular
structure which preserves the relative position of each word and
scales the rectangles they occupy proportionally to word occurrence
probabilities or collocation statistics. Interactive restructuring of
a concordance browser is enabled through the interface. This
restructuring combined with color highlighting of the concordance
lines creates a powerful technique for investigating significant
colocation patterns. The space filling design allows for perceptually
efficient comparison of word collocation statistics such as
mutual information, z-score and log-log score, while preserving some
of the context available on a concordance list. Figure~\ref{fig:mosaic}
shows two examples of concordance mosaics for the word ``regime'',
showing the effect of using a collocation metric to re-scale a full
concordance summary.

Frequency lists enable quick comparisons between words and terms for a
particular corpus, collocation or sub-corpus. In translation studies,
it is common for scholars to compare several lists from different
sub-corpora. In studying translation styles one migh wish, for
example, to compare sub-corpora corresponding to translations by a
given translator, from a given source language, etc. As the frequency
distributions for different sub-corpora will typically vary,
comparisons of words in the same rank order in different lists
sometimes may be misleading. For this reason, modnlp/tec introduced a
tool which enables visual comparisons of frequency lists using slope
lines to connect words on the same relative positions of logarithmic
scaled axes \cite{bib:SheehanMasoodianLuz-avi18} which corrects any
distortions due to differing text sizes.

A less commonly encountered feature of corpus tools is visualisation
of concordances in parallel corpora. While this feature is of great
relevance to translation and translation studies, visualising source
and target translated texts side by side disrupts the visual scanning
of concordance patterns. Perhaps for this reason, this form of display
is not commonly used outside  specialised localisation
tools \cite{bib:LewisEtAl09sfawl}. However, Both Sketch and modnlp
have add-on modules for basic parallel concordancing. 

\subsection{Basic computational linguistics methods}
\label{sec:basic-comp-ling}

Some of the functions of the above described tools, and others like
them, draw mostly on the notion of {\em collocation}
\cite{bib:Sinclair91CorpusConcordance}. Collocations are words that
occur together often enough to be noticed as textual units, being
referred to in the context of translation as {\em terms}. In this
sense, terminology extraction is a task related to translation which
employs computational linguistic methods. In addition to descriptive
methods, such as concordancing, quantitative analytic methods are
employed to determine how often is often enough as a pattern of word
co-occurrence. The basic framework is one of hypothesis testing, where
one wishes to assess whether a given co-occurrence pattern forms a
collocation or is a merely ``accidental'' co-occurrence
\cite{bib:ManningSchutze1999a}. Accidental co-occurrence has been
equated with statistical independence, in a probabilistic
framework. Thus, the null hypothesis in a test of statistical
significance is formulated as the statement that $P(x,y) = P(x)P(y)$
where $P(x,y)$ is the (joint) probability that words $x$ and $y$
appear together in a given context, while $P(x)$ is the (prior)
probability of occurrence of word $x$.  Different statistical tests
have been used in the corpus analysis literature. The most common
approach is to employ the t-test, where the $t$ statistic is computed
to determine the probability that the difference between the expected
co-occurrence distribution (modelled as a Bernoulli distribution) and
the empirically observed distribution occurred by chance. The null
hypothesis is rejected if this probability is smaller than a set
value, usually $0.05$. Other tests have been proposed to account for
the fact that word probabilities are generally not normally
distributed (an implicit assumption of the t test). Among the
non-parametric tests used, the most common is Pearson's $\chi^2$
test. This test is of particular interest in translation as it has
also been employed to identify translation pairs in parallel texts
\cite{bib:GaleChurch91id}.
    
Other statistical tests and information theoretic measures of word
association used in translation and corpus analysis include the
z-score, likelihood ratios, information gain, and mutual
information. The last two can be regarded as information theoretic
measures, as they are motivated by properties of distributions. The
mutual information score\footnote{The version we will describe here is
  also known as ``pointwise mutual information.'' as it deals with
  {\em values} of the random variables that describe word occurences,
  rather than the random variables themselves.} has been widely
employed. It essentially computes the following function of the word
frequencies:
$PMI(x,y)= \log \frac{P(x, y)}{P(x)P(y)}= \log \frac{P(x|y)}{P(x)}$.
PMI thus gives a measure of association; the larger its value, the
greater the dependency between $x$ and $y$ (and the likelihood that
they will form collocations) in the text. This metric is indeed the
basis of the re-scaling of words on the second (bottom) mosaic shown
in Figure~\ref{fig:mosaic}. The words are scaled according to a metric
$M(w,k) = \frac{count(w,k)}{N} \times \frac{\sum_x
  count(x)}{count(w)}$,
which is the ratio of the relative frequency of word $w$ in the
context of word $k$ to the relative frequency of $k$ in corpus. It can
be easily shown that this scaling factor is closely related to mutual
information, that is, $M(w,k) \propto 2^{PMI(k,w)}$.

\subsubsection{Machine learning}

In the last decades, work in computational linguistics has been
increasingly dominated by machine learning approaches. Machine
learning is an umbrella term for a number of methods which aim to
detect patterns and regularities in data by automatic algorithmic
means. According to \citet{bib:Mitchell1997MLbook}, a learner is a
system whose performance ``with respect to some class of tasks $T$ and
performance measure $P$'' improves as the system is exposed to
experience $E$. In the case of computational linguistics, the tasks
range from classification of texts, words or segments, to labelling
and mapping sequences to sequences. Performance measures include
accuracy, precision, recall, and F scores (among other measures), and
the experience (data) corresponds to the corpus itself, or part of it,
presented to the learning algorithm as a formal object we will refer
to as its {\em representation}.

The learning task is usually conceptualised as function approximation:
given a true function $f$, the system must learn an approximation
$\hat f$. The exact nature of the function to be learnt depends on the
task. In a clustering task, for instance, the function to be
approximated could map texts to sets (formally $f: C \to 2^C$,
where $C$ may be, for instance, the set of texts in a corpus), a
classification task might map texts into sets of labels, a
sequence-to-sequence mapping task might map sequences of words to
sequences of parts of speech, or sequences of words in a source
language to sequences of words in a target language, and so on.  

Of the categorisation tasks, the most relevant to translation practice
and studies are word category disambiguation (part-of-speech tagging)
and word sense disambiguation \cite{bib:JurafskyAndMartin08}. In
category disambiguation a word is assigned a grammatical category (a
part of speech tag), for instance, the word ``word'' may be assigned
the category of noun in most contexts or the category of verb in a few
contexts. Corpus analysis tools such as the above described often
allow the user to specify a category for the keyword in order to
restrict the search. Such categories are typically assigned to the
indexed text automatically, through machine learning algorithms. In
sense disambiguation, words or terms are assigned semantic
labels. Although search by word senses is less common in corpus tools,
it is often useful in machine translation, and computer-assisted
translation. Both tasks can be conceptualised, alternatively, as
word-to-category or sequence-to-sequence mappings. Machine learning
algorithms that have been employed to learn word to category mappings
of part of speech or sense tags include maximum entropy
\cite{bib:Ratnaparkhi96} and decision trees
\cite{bib:MarquezRodriguez98p} among others. Sequence to sequence
models include hidden Markov models \cite{bib:CuttingKupiecEtAl92},
conditional random fields \cite{bib:LaffertyMcCallumPereira01crf}, and
more recently, ``deep learning'' models such as bidirectional long
short-term memory have been proposed for multilingual tagging
\cite{bib:PlankSoegaardGoldberg16m}.

Machine translation systems often  make use of the results of
categorisation tasks, and most modern systems incorporate additional
machine learning methods. This is true of both statistical machine
translation \cite{bib:Lopez08smt} and of hybrid approaches which
incorporate corpus and rule-based components
\cite{bib:Costa-JussaFonollosa15l}. Deep learning methods have also
had a significant impact on machine translation performance
\cite{bib:SutskeverVinyalsLe14s} and continue to be an active field of
research \cite{bib:NeubigWatanabe16opsmt}. 

A distinction is sometimes made between supervised and unsupervised
learning, meaning learning tasks where the experience used is guided
by human feedback (often in the form of annotation of the training
data), or solely by features of the data themselves. The above
described tasks might be regarded as supervised tasks. Typical
unsupervised tasks in computational linguistics include clustering of
terms or documents into homogeneous sets. Unsupervised tasks are
commonly used for extraction of input representations for supervised
tasks \cite{bib:EmmsLuz07MLFNLP}, or directly in inference tasks such
as word category and sense disambiguation. In such tasks, it is common
to represent the words occurring in a corpus as a co-occurrence
matrix, as shown in Table~\ref{tab:samplecooc}. A simple clustering of
these terms by the unsupervised k-means algorithm would create
separate clusters for \{``stake''\}, \{``usair'', ``merger'',
``twa''\}, \{``acquire'', ``acquisition''\}, and \{``voting'',
``buyout'', ``ownership'', ...\} which intuitively appear to share
similar semantics. 

\begin{table}[phtb]
  \centering\footnotesize
  \caption{Sample co-occurrence matrix for a subset of REUTERS-21578
    \cite{bib:EmmsLuz07MLFNLP}}
  \label{tab:samplecooc}
  \begin{tabular}{lr@{ }r@{ }r@{ }r@{ }r@{ }r@{ }r@{ }r@{ }r@{ }r@{
      }r@{ }r@{ }r@{ }r@{ }r@{ }r@{ }r@{ }r@{}}                                                                                            \\ \hline
             & \rot{usair} &\rot{voting }&\rot{buyout }&\rot{stake }&\rot{santa }&\rot{merger }&\rot{ownership }&\rot{... }&\rot{manufactures }&\rot{... }&\rot{attractive }&\rot{undisclosed }&\rot{aquisition }&\rot{twa }&\rot{interested} \\  
usair        & 20          & 2      & 0      & 1     & 0     & 4      & 1         & ... & 0            & ... & 0          & 0           & 2          & 14  & 1          \\
voting       & 2           & 10     & 0      & 2     & 0     & 1      & 0         &     & 0            &     & 0          & 0           & 0          & 2   & 0          \\
buyout       & 0           & 0      & 8      & 1     & 0     & 2      & 0         &     & 0            &     & 1          & 0           & 1          & 0   & 0          \\
stake        & 1           & 2      & 1      & 62    & 0     & 0      & 0         &     & 0            &     & 0          & 0           & 0          & 1   & 0          \\
santa        & 0           & 0      & 0      & 0     & 7     & 3      & 0         &     & 0            &     & 0          & 2           & 0          & 0   & 1          \\
merger       & 4           & 1      & 2      & 0     & 3     & 48     & 0         &     & 0            &     & 1          & 0           & 5          & 4   & 3          \\
ownership    & 1           & 0      & 0      & 0     & 0     & 0      & 6         &     & 0            &     & 0          & 0           & 0          & 1   & 0          \\
             & ...         &        &        & ...                                                                                                                      \\
manufactures & 0           & 0      & 0      & 0     & 0     & 0      & 0         &     & 4            &     & 0          & 0           & 0          & 0   & 0          \\
             & ...         &        &        & ...                                                                                                                      \\
attractive   & 0           & 0      & 1      & 0     & 0     & 1      & 0         &     & 0            &     & 4          & 0           & 0          & 0   & 0          \\
undisclosed  & 0           & 0      & 0      & 0     & 2     & 0      & 0         &     & 0            &     & 0          & 24          & 4          & 0   & 0          \\
acquisition  & 2           & 0      & 1      & 0     & 0     & 5      & 0         &     & 0            &     & 0          & 4           & 60         & 2   & 0          \\
twa          & 14          & 2      & 0      & 1     & 0     & 4      & 1         &     & 0            &     & 0          & 0           & 2          & 20  & 1          \\
interested   & 1           & 0      & 0      & 0     & 1     & 3      & 0         &     & 0            &     & 0          & 0           & 0          & 1   & 6          \\
\hline 
  \end{tabular}
\end{table}

\subsection{Distributional semantics}

The example above illustrates the basics of the {\em distributional
  semantics} approach, which has its roots in corpus linguistics and
early machine translation research \cite{bib:Weaver55t}, and has been
influential in computational linguistics applications to translation.

Distributional semantics posits that the semantics of a word is
determined by its {\em context}, that is, the words that occur in its
vicinity. An early computational realisation of this idea is the {\em
  vector space model}, commonly used in information retrieval
\cite{bib:SaltonWongYang75,bib:BaezaYatesRibeiroNeto99} where a
textual entity such as a word, a document or a text segment is
represented as a vector in a common algebraic model. Each row (or,
equivalently, each column) in the co-occurrence matrix of
Table~\ref{tab:samplecooc}, for instance, can be regarded as a vector
representation of a word. Thus, the word ``usair'' may be
represented by vector $v = (20, 2, 0, 1, \ldots)$, where each value
indicates the number of times ``usair'' occurs in the same context as
other words in the lexicon. This type of
representation is also known as {\em bag of words} in text
categorisation \cite{bib:Sebastiani2002}. 

\begin{figure}[htbp]
  \centering
  \includegraphics[width=\linewidth]{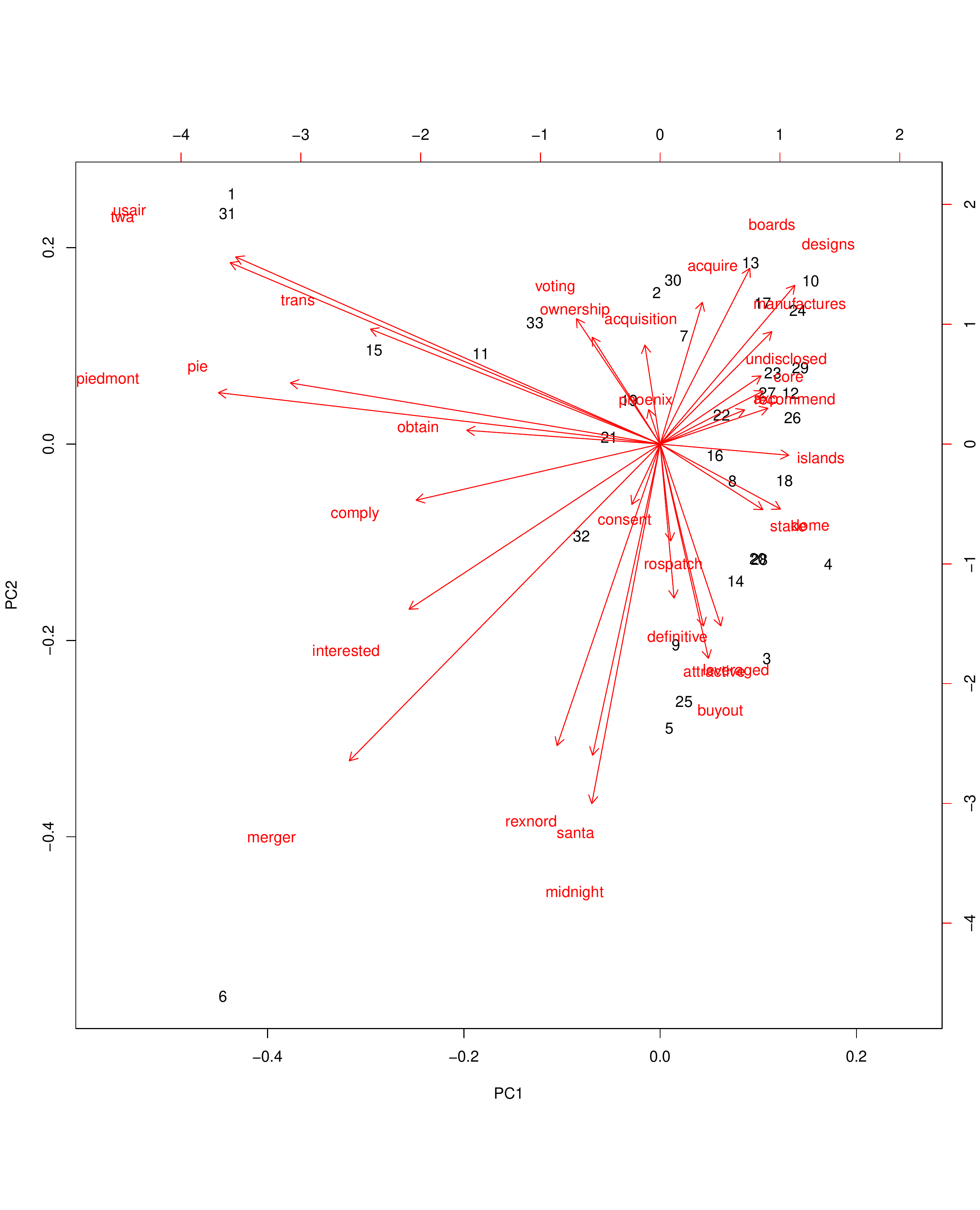}
  \caption{Projection of a set of REUTERS-21578 words
    (Table~\ref{tab:samplecooc}) onto two dimensions, by principal
    component analysis.}
  \label{fig:pcaprojection}
\end{figure}

The vector space model has served as a basis for a number of feature
extraction methods. Feature extraction is the process of reducing the
dimensionality of the vector space by projecting it onto a small
number of dimensions. Each projection onto a lower dimensional space
is regarded, in this framework, as a semantic representation of a
word. Different methods exist which compute such projections. The most
commonly used methods are principal component analysis (PCA), based on
eigenvalue decomposition of the covariance matrix of the vector space,
and latent semantic indexing (LSI), which is based on a generalisation
of eigenvalue decomposition known as singular value
decomposition. \citet[ch. 15]{bib:ManningSchutze1999a} describe the
use of LSI to project a high dimensional vector space of words onto a
reduced space of independent dimensions by applying singular value
decomposition to the original document matrix. The intuitive appeal of
these methods can be illustrated by a simple projection of the word
vectors of Table~\ref{tab:samplecooc} onto the vector space's two
dimensions (defined by the first two principal components), as given
by singular value decomposition
(Figure~\ref{fig:pcaprojection}). While there is considerable
sparsity, it can be seen that the model succeeds in placing a number
of semantically related words in close proximity on the 2-dimensional
plane (e.g. ``usair'' and ``twa'' are both airlines, the vectors for
``voting'', ``ownership'', and ``acquisition'' are in close proximity,
and so on).

A recent development in this field is the combination of the vector
space model with neural networks for language modelling
\cite{bib:TurianRatinovBengio10w}. A {\em language model} is a
probabilistic model which encodes the probability of occurrence of a
word in a context, most often for predicting the next work given a
sequence of words. In neural network implementations of such models,
the network's inputs are word vectors and the learning task is to
determine optimal parameter for the model (the word sequence
probabilities). The input vectors are mapped onto a lower dimensional
space through a network layer called the {\em embedding layer}. In
addition to providing compact and effective representations for
machine, these embeddings have interesting semantic properties from a
distributional semantic perspective. In an often cited example,
\citet{bib:MikolovSutskeverEtAl13d} show how the vector
representations for ``king'', ``queen'', ``man'' and ``woman'', learnt
from a corpus, can be combined through algebraic operations so that
``king - man + woman'' results in a vector that is very near the
vector for ``queen'' in the lower dimensional space yielded by the
embedding.  Embedding techniques have since been decoupled from their
original application to language modelling and become an area of
research in itself, with application in several areas of computational
linguistics, including translation
\cite{bib:AlmeidaXexeo19wem}. Similar techniques have also been
applied to produce cross-lingual representations from aligned words
\cite{bib:LuongPhamManning15b} or sentences
\cite{bib:HermannBlunsom13m}.

\section{Contexts of application}
\label{sec:contexts}

Contexts of application for computational linguistic methods include,
besides stand-alone machine translation systems, the use of corpora
for translator training and education
\cite{bib:ZanettinBernardiniStewart14c}, the use of comparable and
parallel texts, concordancing and machine translation tools in
translation practice \cite{bib:DohertyKaramanisLuzCSCW2012}, the use
of corpora and text visualisation tools in the analysis of the
linguistic behaviour of translators
\cite{bib:LuzCTSBook2011,bib:BakerIJCL99}, and the use of machine
learning and text categorisation methods for the characterisation of
translationese \cite{bib:BaroniBernardini05,bib:IliseiInkpenEtAl10id},
as mentioned above.

Machine translation has been an important area of application for many
of the above described techniques and tools. While rule-based machine
translation systems have benefited from these tools and techniques, it
is in the area of statistical machine translation that they have had
the greatest impact. As mentioned, methods such as word category and
sense disambiguation are often incorporated into these systems to
improve performance \cite{bib:CarpuatWu07im}, and conversely new
developments in neural network based translation have recently being
used in disambiguation tasks \cite{bib:GonzalesMascarellSennrich17im}.
As pointed out earlier, new computational linguistics methods such as
cross-lingual embeddings, which can be induced automatically from
parallel (and sometimes from comparable) corpora have also contributed
semantically compelling representations of terms across language
pairs.

Machine translation has been increasingly used for ``gisting''
(understanding the essential message of a text) both in informal
multi-lingual communication and in commercial settings
\cite{bib:KoponenSalmi15}. Machine translation has also been
incorporated into the work of professional translators and
post-editors, often with unanticipated implications to their
established work practices
\cite{bib:KaramanisLuzDoherty11tpw,bib:MoorkensOBrien17as} and their
collaborations with colleagues in the workplace
\cite{bib:DohertyKaramanisLuzCSCW2012}.

Corpus management tools have also been extensively used in commercial
translation settings, more commonly so than machine translation. The use
of concordances as a means of selecting a suitable translation among a
set of alternatives is now a feature of most ``localisation'' (i.e. the
process of translating, most often software, manuals and websites, and
adapting their contents to a local region, country or culture) systems
\cite{bib:LewisEtAl09sfawl,bib:MoorkensOBrien17as}. Corpora and corpus
software also play an important role in translator education. Tools
such as concordance browsers, statistical analysis packages,
and visualisation software form an important part of practical training.
Technologies such as machine translation are also being incorporated
into the curriculum
\cite{bib:DohertyKenny14}. See articles in the collection edited by
\cite{bib:ZanettinBernardiniStewart14c} for different perspectives and
approaches in this area.

Finally, it is indisputable that the discipline of corpus based
translation studies has its methodological foundations firmly based on
the capabilities afforded by computational processing of large volumes
of text. Work in this discipline was strongly influenced by the work
of John Sinclair
\cite{bib:Sinclair03Reading,bib:Sinclair91CorpusConcordance}, and the
analyses based on the translational English corpus through the use of
the TEC corpus tools \cite{bib:LuzCTSBook2011} greatly contributed to
the methodology of some of the field's seminal work
\cite{bib:Bakerothers93c,bib:BakerIJCL99,bib:Baker04}. Computational
linguistics methods that rely strongly on machine learning are used
less often in translation studies, but the work of
\citet{bib:BaroniBernardini05} and \cite{bib:IliseiInkpenEtAl10id}, in
which the authors employed machine learning model to identify
characteristic features of translated text, are interesting
exceptions.

\section{Critical issues and topics}
\label{sec:crit-issu-topics}

As computational linguistics and machine learning methods make further
inroads into translation and translation studies, potential issues
arise which need to be examined from a methodological
perspective. These issues include: the need to support the translator
and the translation scholar with tools and user interfaces that enable
usable and effective access to large volumes of text and facilitate
selection and comparison of sub-corpora, the interpretability of the
statistical and connectionist models employed, and issues concerning
the generation and validation of hypotheses and conclusions reached
through the use of computational linguistics methods.

Clearly, good user interface support for users of computational
linguistics tools is crucial in applications such as translator
training and computer assisted translation. With the advent of corpus
based translation studies, it has also become important to provide the
research community with usable tools which will also allow researchers
to document and share their work. Tools such as the Sketch Engine
\cite{bib:KilgarriffBaisaEtAl14sen}, the BYU corpora
\cite{bib:DaviesFuchs15exwen,bib:Davies10m}, the CQPweb corpus
analysis system \cite{bib:Hardie12c}, and the modnlp/tec
\cite{bib:LuzCTSBook2011} software suite can be seen as efforts
towards these goals. However, the use of analytic
tools and corpora is still hampered by software access and licensing
constraints. With the exception of modnlp/tec, all of the above
mentioned tools are exclusively web based. While using the web as a
platform certainly facilitates access, the absence of a stand-alone,
offline tool limits the more experienced user's flexibility and
their ability to explore their own resources, which they might not wish or
have the legal right to share. Another issue of access concerns
licensing terms which might limit access to software source code, or
prevent it entirely. Unfortunately, most web based tools in this area
are commercial products, provide very limited functionality, or
charge fees for ``premium access''. Two exceptions among the tools
cited are CQPweb and modnlp/tec, both of which are distributed under
FLOSS licenses.  The ability to inspect and modify source
code has been increasingly regarded as a crucial aspect of
reproducibility in data-intensive research \cite{bib:Hutson18ar}.  If
corpus based studies are to develop a robust methodology for the use of
computational linguistics tools and methods, the issue of sharing
source code as well as data needs to be addressed. A related issue is
the availability and stability of the software. Users of purely
web based tools are entirely dependent on the software provider for
their analytic work. If the tool's underlying algorithms changes, or
the tool is withdrawn from public access, the corpus scholar
potentially faces the situation of having their analyses invalidated
(in the case of algorithm changes) or uncorroborated (in the case of
access withdrawal). While several web-based text visualisation tools
have appeared recently\footnote{See for instance, \citet{bib:VoyantTools21} which generates interactive word clouds and summary statistics from text uploaded by the user or harvested from the web, and \cite{bib:TextTexture21} which renders texts as visually appealing graphs.}, these 

Ideally, standardised, FLOSS platforms will
be built in the future which allow corpus and translation researchers
to document and share their analyses, perhaps along the lines of what
has been done for ``vernacular visualisation'' 
\cite{bib:ViegasWattenberg08tt}, where users are encouraged to
produce, document and share their analyses. 

As machine learning models and algorithms start to become part of the
translator's and translator scholar's toolbox, the issue arises of
ensuring that these models and algorithms are well understood and
properly used. This is not a trivial issue, witness the concerns in
other traditionally data-intensive research communities regarding the
misuses and misinterpretations of simple statistical tests. As machine
learning models tend to be a lot more complex, and in many cases more
opaque than the models used in traditional statistical testing, their
potential misuse should be a cause for concern. For instance, while
the interpretation of a co-occurrence matrix is straightforward, an
embedding vector which might involve non-linear transformations of the
original data would not be as easy to interpret. Therefore, somewhat
in opposition to the need to provide usable interfaces just discussed,
it seems that ensuring the proper use of computational linguistics
methods based on machine learning will require a level of proficiency
in these methods which cannot be achieved simply through better user
interfaces. This is however an active area of research, and new techniques
of visualisation and model explanation are being proposed which might
alleviate this problem in future. A related issue concerns the
incorporation of new technology into translator
education. \citet{bib:DohertyKenny14} argue for ways of incorporating
statistical machine translation into the translation training
curriculum in a way that promotes a good understanding of the
underlying technology and therefore empowers the translator. 

From education to scholarly studies, it is clear that the arrival of
new language technologies has disrupted established practices, so that
emerging fields such as corpus-based translation studies now find
themselves at a stage where further progress in methodology can only
be made through the joint efforts of researchers from several
disciplines, including translation scholars, linguists, statisticians
and computer scientists. This state of affairs has been anticipated to
some extent \cite{bib:Baker00t}, but as the field evolves the need for
interdisciplinary collaboration becomes more evident. 

\section{Recommendations for practice}
\label{sec:recomm-pract}

Although there are several implementations of computational
linguistics tools, many of which have been released as FLOSS software,
which could in principle be used in translation and translation
studies projects, integrating such tools into usable and coherent tool
sets for use by translator scholars can be challenging.  Most of these
tools require the user to have at least basic programming skills,
though the requirements are often considerably higher. Access to
suitable corpora and other data resources is also often an
issue. Fortunately, however, there have been developments and efforts
towards standardisation and methodological consolidation
\cite{bib:Zanettin14t}, which may ease the burden of translators and
translation scholars wishing to use computational linguistic tools.

Based on the developments, tools and methodological challenges
discussed in this chapter so far, we can now summarise some
recommendations for practice regarding the use of computational
linguistics in translation work and research:

\begin{itemize}
\item In undertaking corpus-based work, consider the
  accessibility and potential restrictions on sharing of linguistic
  material. Acquisition and use of such resources is often a
  demanding and time-consuming first step in both research and
  commercial translation.
\item Choose usable computing tools which give you the necessary
  flexibility with regards to use and management of your own corpora,
  in addition to online corpora, and which allow
  the user to progress naturally from simpler to more complex modes of
  analysis.
\item Learn the basic underlying theory of the computational
  linguistics methods used in your analysis or work, in order to
  ensure that these methods are adequately used. This is important in
  any data analysis task, but specially important in the analysis of
  translation, where data sparsity and complex, 'black box' models
  often conspire to produce invalid results. Learning basic
  programming skills may also save the corpus and translation scholar
  a lot of time and frustration.
\item Consider reproducibility of research results. Whenever possible,
  prefer FLOSS software which allows other researchers to closely
  inspect your methods and replicate your results.  Prefer open,
  well-documented standards for text encoding, storage, and access.
\item Monitor the literature for new developments in text
  visualisation technology. This is a fast-moving field which is
  gaining increasing importance with the widespread use of neural
  networks and other machine learning models in computational
  linguistics. Text visualisation tools can act as an effective
  complement to computatinal linguistics methods. For the translation
  researcher, this is important as visualisation tool may help the
  user gain an accurate overview of the data and formulate research
  hypotheses. For the practitioner, these tools have the potential to
  improve user interface, and the user's overall experience of the
  translation process.
\end{itemize}

\section{Further reading}
\label{sec:further-reading}

The references in this chapter provide a comprehensive bibliography of
computational linguistics tools and methods of relevance to
translation and translation studies. However, a few key reference
texts may help guide the reader through specific methods and
approaches in computational
linguistics. \citet{bib:ManningSchutze1999a}, ``Foundations of
Statistical Natural Language Processing'' provides a good overview of
the basic methods in computational linguistics. The book by
\citet{bib:JurafskyAndMartin08} describes a large number of techniques
used in natural language processing, including knowledge based
methods. \citet{bib:Mitchell1997MLbook} is a very readable
introduction to machine learning. \citet{bib:EmmsLuz07MLFNLP} provide
a hands-on tutorial on the use of machine learning in language
analysis. \citet{bib:Lopez08smt} surveys statistical machine
translation technology, \citet{bib:NeubigWatanabe16opsmt} reviews
novel optimisation techniques in this field, and
\citet{bib:Costa-JussaFonollosa15l} surveys the latest trends in
systems that combine rule-based and statistical methods in machine
translation. The paper by \citet{bib:Hardie12c}, contains a brief
survey of web-based corpus analysis tools, and \citet{bib:Zanettin14t}
offers a comprehensive introduction to corpus tools and methods for
translation scholars and practitioners. Finally, new developments in
vector representations of text, such as word embeddings are concisely
reviewed by \citet{bib:AlmeidaXexeo19wem}, and an extensive survey of
cross-lingual models is presented by \cite{bib:RuderVulicSoegaard17}.
\cite{bib:CaoCui16in}\cite{bib:OBrien12t}

\section{Related topics}
\label{sec:related-topics}

Topics related to applications of computational linguistics to the
practice and study of translation include: methods for visualisation
of text and language models, studies of translator work and
collaboration in translation, and the relation between computational
linguistics and qualitative methods.

\bibliography{luz,biblio,lusz-outline}
\bibliographystyle{chicago-sl}


\end{document}